\documentclass[10pt,twocolumn,letterpaper]{article}
\usepackage{multirow}  
\usepackage{iccv}              % To produce the CAMERA-READY version
\usepackage{booktabs}
\usepackage{colortbl} 
\usepackage[dvipsnames]{xcolor} % 使用 table 选项支持表格内着色
\usepackage{capt-of} 
\usepackage{graphicx}
\definecolor{lightblue}{RGB}{232, 244, 255}
\definecolor{paleRed}{RGB}{255,204,204}
\definecolor{palePurple}{RGB}{210,200,250}

\definecolor{iccvblue}{rgb}{0.21,0.49,0.74}
\usepackage[pagebackref,breaklinks,colorlinks,allcolors=iccvblue]{hyperref}

\def\paperID{1845} % *** Enter the Paper ID here
\def\confName{ICCV}
\def\confYear{2025}

\title{RealGeneral: Unifying Visual Generation via Temporal In-Context Learning with Video Models}

\author{Yijing Lin, Mengqi Huang, Shuhan Zhuang, Zhendong Mao\textsuperscript{†} \\
University of Science and Technology of China\\
{\tt\small \{lyijing, huangmq, zhuangsh\}@mail.ustc.edu.cn, zdmao@ustc.edu.cn}
}

\begin{document}

\twocolumn[{
    \renewcommand\twocolumn[1][]{#1}
    \maketitle
    % 可以视情况调整垂直距离
    \vspace{-1.46em}
    \begin{center}
        \includegraphics[width=1\linewidth]{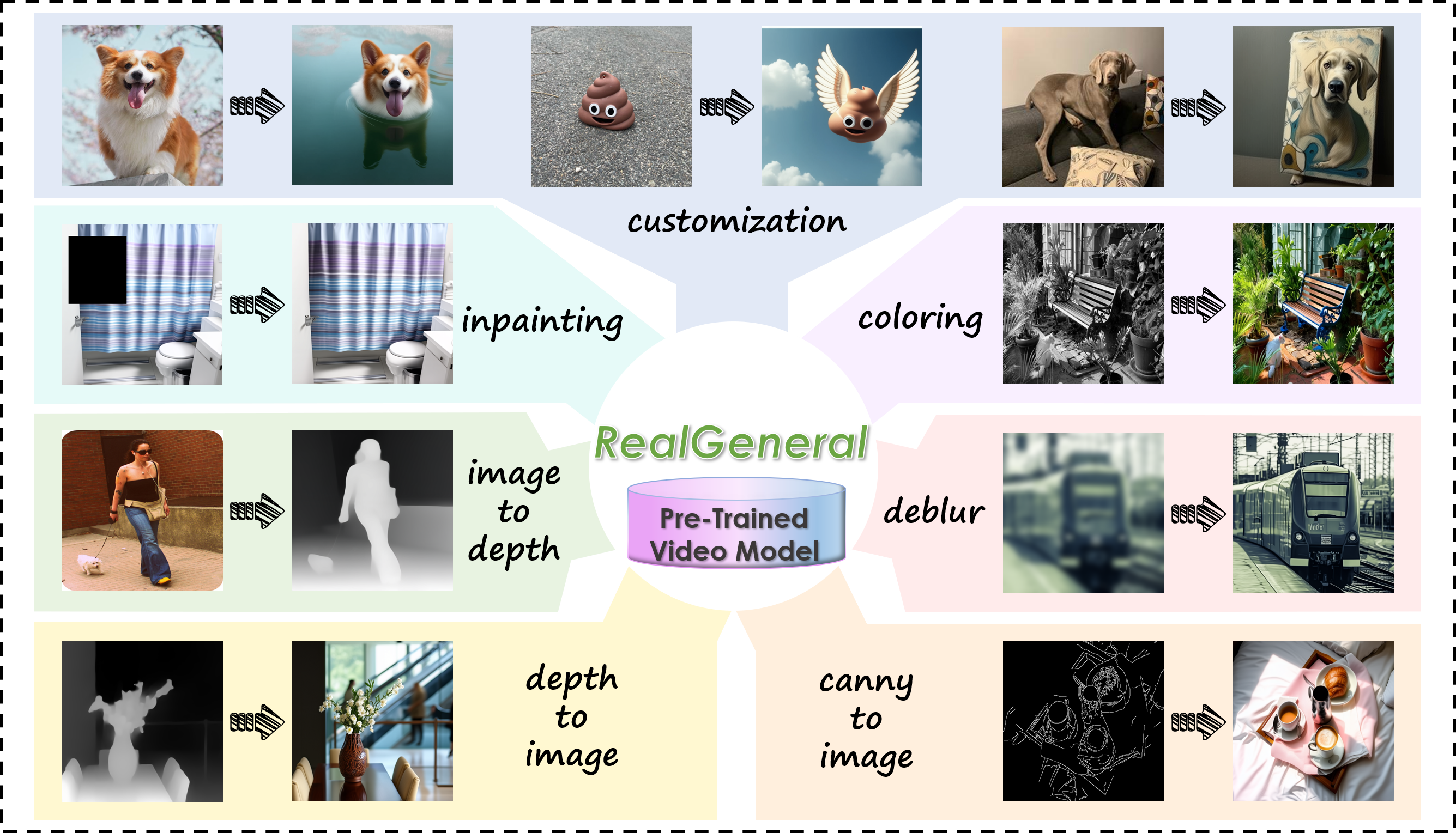}
        \vspace{-1.35em}
        \captionof{figure}{Results from RealGeneral, demonstrate the ability to produce high-quality images from diverse input conditions.}
        \label{fig:result}
    \end{center}
    % \vspace{1em}
}]
\begin{abstract}
\let\thefootnote\relax\footnote{\textsuperscript{†}Zhendong Mao is the corresponding author.}
Unifying diverse image generation tasks within a single framework remains a fundamental challenge in visual generation. While large language models (LLMs) achieve unification through task-agnostic data and generation, existing visual generation models fail to meet these principles. Current approaches either rely on per-task datasets and large-scale training or adapt pre-trained image models with task-specific modifications, limiting their generalizability. In this work, we explore video models as a foundation for unified image generation, leveraging their inherent ability to model temporal correlations. We introduce RealGeneral, a novel framework that reformulates image generation as a conditional frame prediction task, analogous to in-context learning in LLMs. To bridge the gap between video models and condition-image pairs, we propose (1) a Unified Conditional Embedding module for multi-modal alignment and (2) a Unified Stream DiT Block with decoupled adaptive LayerNorm and attention mask to mitigate cross-modal interference. RealGeneral demonstrates effectiveness in multiple important visual generation tasks, \eg, it achieves a 14.5\% improvement in subject similarity for customized generation and a 10\% enhancement in image quality for canny-to-image task. Project Page: \href{https://lyne1.github.io/realgeneral_web/}{realgeneral\_web}; GitHub Link: \href{https://github.com/Lyne1/RealGeneral}{https://github.com/Lyne1/RealGeneral}
\end{abstract}    
\section{Introduction}
\label{sec:intro}
With the rapid development of generative modeling~\cite{ho2020denoising,ho2022cascaded,huang2022dse,chang2023muse,huang2023towards,xu2024prompt,liu2025one}, image generation has gained considerable attention in various application domains. Although foundation models such as Stable Diffusion~\cite{esser2021taming, podell2023sdxl} and FLUX~\cite{flux2024} have achieved remarkable success in unimodal text-to-image generation, processing diverse inputs (\eg, text, depth maps, conditional images) and generating corresponding images within a unified framework remains a significant challenge. While large language models (LLMs) have unified text generation~\cite{brown2020language, wei2022chain, achiam2023gpt}, an analogous unified model for the visual domain is still lacking.  
% The primary difficulty lies in integrating multi-modal conditional semantics and distinguishing between visual conditions and generation semantics.

\begin{figure}
    \centering
    \includegraphics[width=1\linewidth]{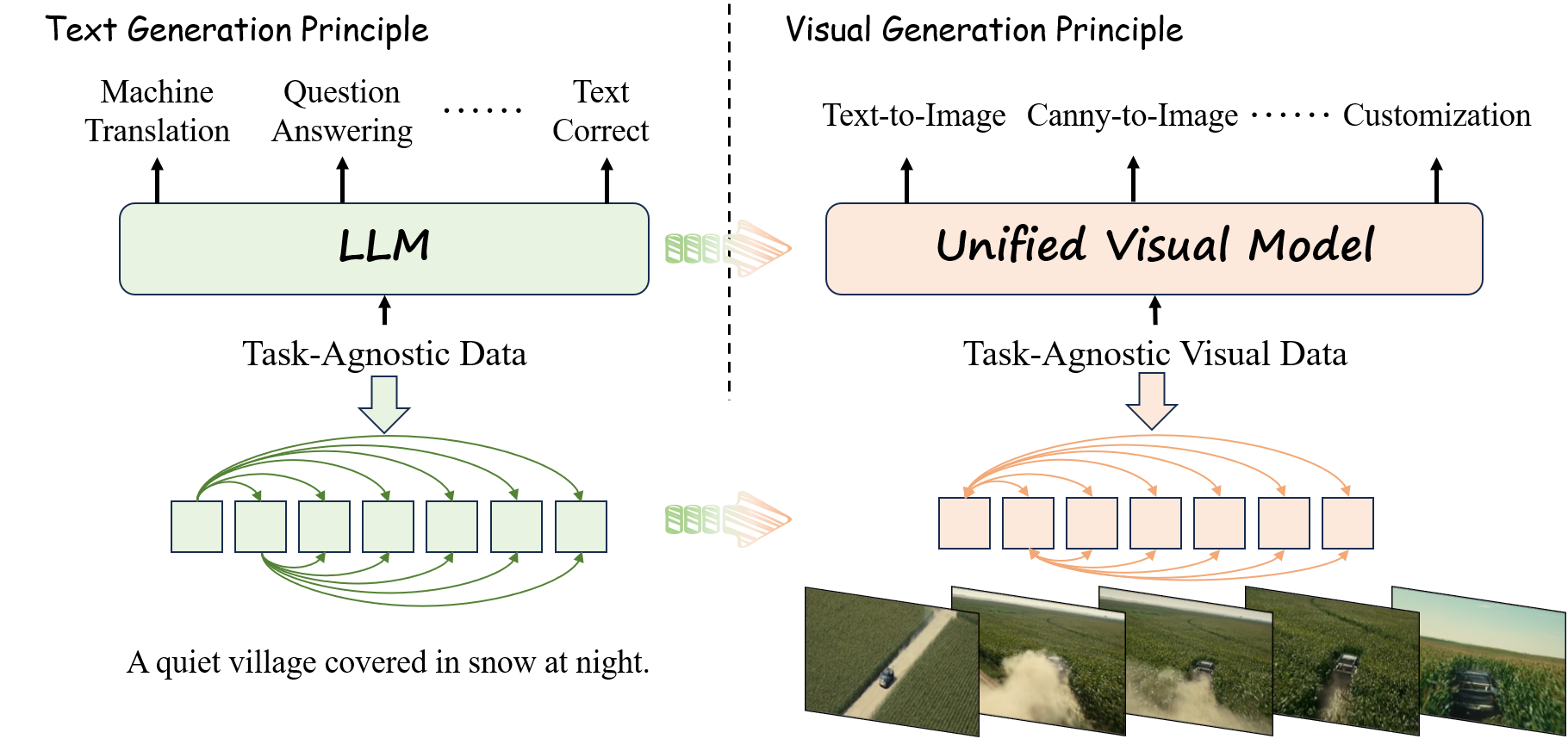}
    \caption{LLMs unify text generation through task-agnostic data and task-agnostic generation. Therefore, unified visual generation models should adhere to similar principles through visual task-agnostic data and task-agnostic visual generation principles.}
    \label{motivation_pic}
\end{figure}

Upon closer examination of the core principles underlying LLMs' success as a unified text generation architecture, we identify two fundamental pillars, \emph{i.e.}, (1) a task-agnostic data principle during pre-training that facilitates the utilization of web-scale corpora for general linguistic knowledge acquisition, and (2) a task-agnostic generation principle that coherently unifies all kinds of textual tasks through next-token prediction. 
Analogously, a promising and effective architecture for unified visual generation should also exhibit (1) \textbf{a task-agnostic visual data principle} for general visual knowledge acquisition, and (2) \textbf{a task-agnostic visual generation principle} that could unify all kinds of multi-modal visual tasks (\emph{e.g.}, customization generation, canny-to-image generation, etc.), as shown in \cref{motivation_pic}.

Recent research on unified visual generation models mainly evolves along two streams, \emph{i.e.}, (1) the training-from-scratch stream~\cite{qin2023unicontrol, wang2023images, bai2024sequential, sun2024x, lin2024pixwizard}, which curates per-task datasets and trains unified models through \emph{de novo optimization}, and (2) the image-model-adaptation stream~\cite{chen2024omnicreator, yu2024anyedit}, which reformulates generation tasks as \emph{partial image synthesis} by leveraging \emph{pre-trained generative priors} (\emph{e.g.}, FLUX). The first stream usually requires vast amounts of high-quality data and massive computational resources~\cite{wang2024lavin, chen2024unireal}. For example, OmniGen\cite{xiao2024omnigen} constructs a dataset of 0.1 billion images for all tasks and trains a unified model using 104 A800 GPUs. The second stream leverages the rich, general representations of pre-trained image generative models by fine-tuning them for specific tasks, thereby reducing computational costs. Some work~\cite{zhang2023adding, ye2023ip, xiao2024omnigen} designs additional condition encoders or task-specific adapters to handle new condition types, while other work~\cite{tan2024ominicontrol} revises only certain modules in the pre-trained models to adapt them to new tasks. Though much progress has been made, both of the streams fail to meet either the task-agnostic visual data principle or the task-agnostic visual generation principle. 
To be specific, the first stream heavily relies on the image dataset, inherently violating the task-agnostic visual data principle. Analogous to LLMs that learn by modeling relationships between sequential tokens, only video data inherently captures inter-frame relationships essential for general visual knowledge acquisition, while single images lack this attribution. Moreover, the second stream not only depends on image data but also adopts specialized architecture for certain tasks, violating both principles.

To address the limitations of existing methods, we further explore whether large-scale pre-trained video generation models can overcome these challenges and provide a more robust framework for unified visual generation. In this study, we posit that such models, which are structurally endowed with both short-term micro-coherence and long-term macro-consistency, serve as a more effective and general foundation for this task.
On the one hand, short-term micro-coherence captures fine-grained temporal dynamics, ensuring precise local details. On the other hand, long-term macro-consistency inherently preserves global structure and semantic consistency across entire sequences. Though great potential, how to effectively harness the pre-trained video models' inherent consistent generation capabilities for unified visual generation remains unexplored.

Building on this insight, we argue that video models, with their inherent capability to model temporal correlations across frames, provide a more natural foundation for unified visual generation. Inspired by in-context learning in LLMs, we reformulate image generation as a conditional frame prediction task. In this formulation, input images are treated as preceding frames in a video sequence, analogous to previous tokens in LLMs, and generate the subsequent frame as the target output. This unified formulation recasts diverse visual generation tasks as instances of conditional frame prediction, eliminating the need for task-specific architectural modifications and thereby satisfying the task-agnostic visual generation principle. Furthermore, video models are trained on task-agnostic data (\ie, videos), adhering to the task-agnostic visual data principle. Therefore, the pre-trained video model only needs a small amount of corresponding task data for efficient fine-tuning, which can elicit the ability to handle various visual generation tasks, analogous to in-context learning in LLMs.

In this study, we propose a new framework termed \textit{RealGeneral}, adapting video diffusion models to image generation by proposing a minimalist framework centered on one-condition-image to one-target-image synthesis. In visual generative tasks, effectively fusing multi-modal inputs without conflating them with the generated image is challenging. To address these challenges, we introduce two core innovations. First, the \textit{Unified Conditional Embedding} module fuses multi-modal inputs, aligning multi-modal conditions through both inter-modal conditional semantic alignment and intra-modal generative semantic distinction. Second, the \textit{Unified Stream DiT Block}, a redesigned video model transformer block that incorporates (1) triple-branch adaLN to decouple feature modulation among text, condition frames, and target frames, eliminating the interference between the condition frame and target frame, and (2) attention mask that prevents text-to-condition interactions while preserving full condition-to-target visual attention to avoid the blending of textual and visual information. 

Our main contributions are summarized as follows:

\textbf{Conceptual contribution.}  We point out two crucial principles underlying the unification of visual generation. Furthermore, we propose a novel formulation that recasts multi-condition image generation as temporal in-context learning within video models, thereby establishing a unified framework for image generation.

\textbf{Technical contribution.} (1) We propose \textit{Unified Conditional Embedding} module for aligning multi-modal inputs. (2) We propose \textit{Unified Stream DiT Block} with decoupled adaLN and attention mask to mitigate interference between different modalities. 

\textbf{Experimental contribution.} We validate our framework across multiple tasks. For instance, RealGeneral outperforms existing methods in the customized generation, achieving a 14.5\% improvement in subject similarity compared to state-of-the-art models. In the canny-to-image task, RealGeneral enhances image quality by 10\%, while in the depth-to-image task, it attains image quality comparable to existing models. Furthermore, training RealGeneral requires significantly fewer computational resources and less data than both unified models and specialized models.

% DINO 14.5%, CLIP-I 6.8%
% FID 6.8%, SSIM 10%
\section{Related work}
\label{sec:formatting}
\subsection{Image Generation}
Image generation has rapidly advanced by combining NLP techniques with transformer architectures. Autoregressive models utilize transformers to predict sequences of discrete codebook codes~\cite{esser2021taming,ramesh2021zero,sun2023emu}. Meanwhile, continuous diffusion models have emerged as a powerful framework for image generation~\cite{ho2020denoising,rombach2022high,dhariwal2021diffusion}.  The integration of large-scale transformer architectures has led to the development of advanced models, such as DiT~\cite{peebles2023scalable, flux2024}. Many efforts have been made to extend the capabilities of generation models. For example, T2I-Adapter~\cite{mou2024t2i} introduces additional encoding layers to improve controllability, while RealCustom~\cite{huang2024realcustom,mao2024realcustom++} disentangles subject similarity from textual controllability by restricting subject influence to semantically relevant regions through a train-inference decoupled framework. However, these methods are designed for specific tasks, enhancing the capabilities of generative models through architectural modifications.

\subsection{Video Generation}
The field of text-to-video models has witnessed remarkable progress. Early efforts to pre-train and scale Transformers for generating videos from text, such as CogVideo~\cite{hong2022cogvideo} and Phenaki~\cite{villegas2022phenaki}, demonstrate significant potential. At the same time, diffusion models have recently made groundbreaking strides in video generation~\cite{li2022efficient,singer2022make,ho2022video,blattmann2023align,guo2023animatediff}. With the introduction of DiT, text-to-video generation has reached a new level, as highlighted by the impressive performances of Sora~\cite{openai2024sora} and CogVideoX~\cite{yang2024cogvideox}.

\subsection{Unified Image Generation}
Current research on unified image generation can be mainly categorized into two streams: training-from-scratch, and image-model-adaptation. Many works focus on the first stream~\cite{qin2023unicontrol, wang2023images, mizrahi20234m, bai2024sequential, sun2024x, chen2024unireal, han2024ace}. For example, PixWizard~\cite{lin2024pixwizard} employs a flow-based DiT~\cite{peebles2023scalable} model that gradually injects conditional and textual information during cross-attention. OmniGen~\cite{xiao2024omnigen} jointly models text and images, applying causal attention to each token in the sequence and bidirectional attention within each image sequence. The image-model-adaptation stream leverages pre-trained image models to improve efficiency. For instance, ControlNet~\cite{zhang2023adding} uses trainable copies and zero convolution layers, enabling Stable Diffusion~\cite{esser2021taming,podell2023sdxl} to adapt specific tasks. OminiControl~\cite{tan2024ominicontrol} builds on FLUX.1~\cite{flux2024}, proposing a multi-modal attention mechanism that facilitates direct interaction between conditions and images. 

Our concurrent work, UniReal~\cite{chen2024unireal}, also reformulates images as discrete frames within a video generation framework. However, there are three critical distinctions between UniReal and our approach. First, the motivations differ: UniReal adopts the training-from-scratch paradigm, only modeling the relationship between discrete frames, whereas RealGeneral introduces two visual principles underlying the unification of visual generation, recasting multi-condition image generation as temporal in-context learning within video models. Second, the design choices diverge: UniReal relies on complex prompt engineering, while RealGeneral achieves controllability without such designs. Third, RealGeneral is significantly more efficient, fine-tuning a pre-trained video model using LoRA with only 0.1\% of UniReal’s 360 million training samples, resulting in substantially reduced computational cost.
\begin{figure*}[t!]
    \centering
    \includegraphics[width=1\linewidth]{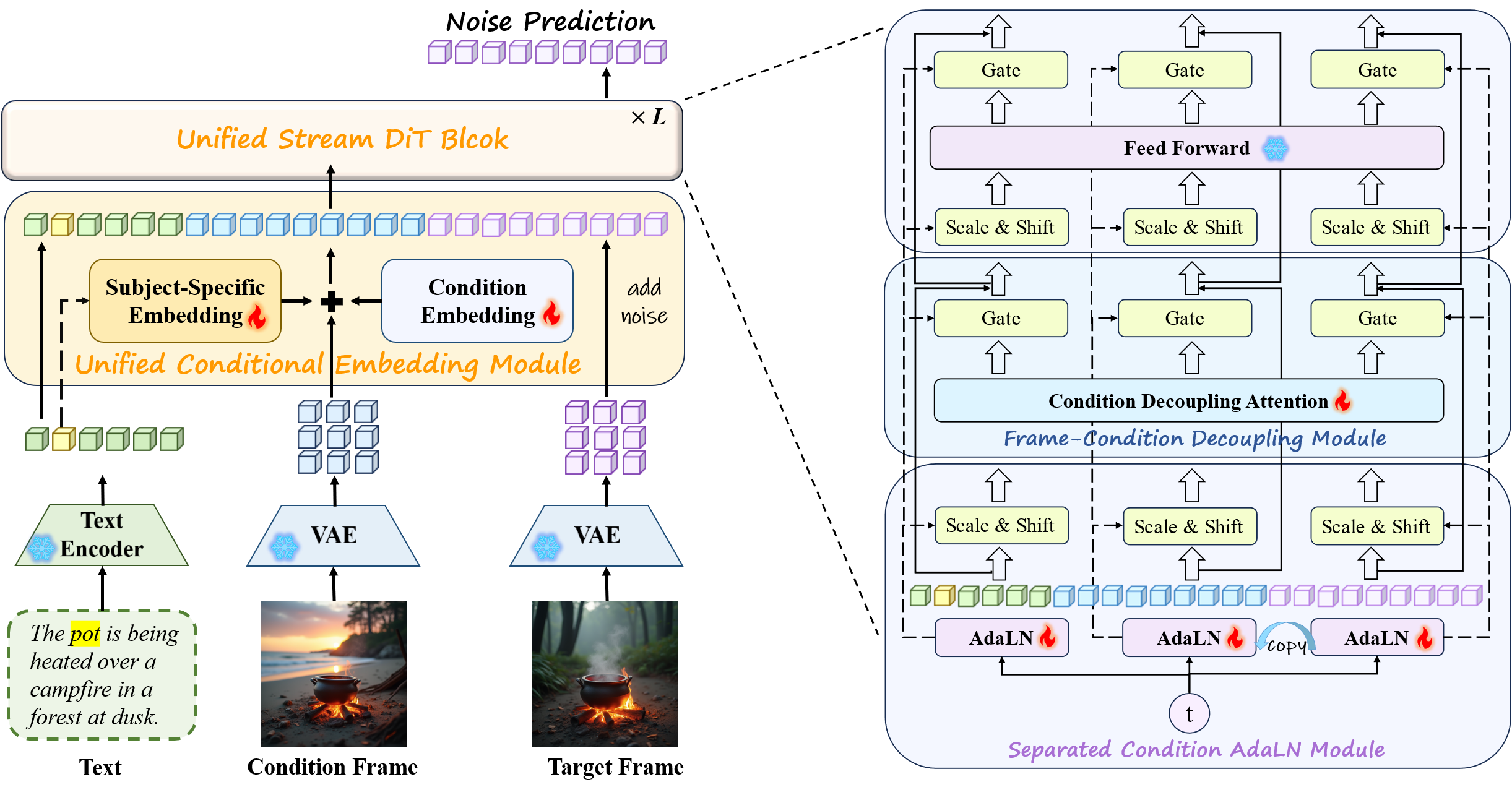}
    \caption{The overview framework of RealGeneral, recasts the image conditional generation task as a next-frame prediction. We show two frames here for simplicity. First, two images are separately encoded by VAE. The Unified Condition Embedding Module integrates textual and global task priors into the condition frame while adding noise to the target frame. Then all tokens are concatenated into a sequence entering the Unified Stream DiT Block. The Separated Condition AdaLN Module modulates text, condition, and target tokens independently via three distinct branches. The Frame-Condition Decoupling Module prevents confusion between text and condition information.}
    \label{framework}
\end{figure*}

\section{Methodology}
Inspired by LLMs that unify textual generation through unlabeled large-scale text pretraining and in-context learning abilities, we argue that the video foundation model pre-trained on massive continuous visual data can similarly unify the visual generation tasks. Analogous to how LLMs autoregressively predict tokens conditioned on previous context, video models can naturally extend this concept to temporal visual frames, predicting subsequent frames from preceding ones. Specifically, given previous frames as the condition frames \{$\mathbf{c}_1$, $\mathbf{c}_2$, $\mathbf{c}_3$ ......\}, video models predicts subsequent frames \{$\mathbf{x}_1$, $\mathbf{x}_2$, $\mathbf{x}_3$ ......\}, by exploiting is temporal modeling capabilities. 

In this work, we focus on a basic two-frame setting: the condition image is considered as the first frame, while the target image (accompanied by descriptive text) forms the second frame, as illustrated in \cref{framework}. In this section, we first introduce our base model in \cref{sec3.1} and then describe three key modules in detail. \cref{sec3.2} details \textit{Unified Conditional Embedding} (UCE) module for aligning multi-modal inputs. \cref{sec3.3} explains \textit{Separated Condition AdaLN} (SC-AdaLN) module enforcing semantic separation between condition and target frames. And \cref{sec3.4} describes \textit{Frame-Condition Decoupling} (FCD) module that employs an attention mask to prevent interactions between the condition image and text. Finally, we describe our task-specific LoRA in \cref{sec3.5} 

\subsection{Preliminaries}
\label{sec3.1}
Our proposed model is based on CogVideoX1.5~\cite{yang2024cogvideox}, a video generation model comprising three core components: (1) a T5 text encoder~\cite{raffel2020exploring} for text processing, (2) a 3D causal VAE for spatio-temporal compression, and (3) a transformer based on the DiT~\cite{peebles2023scalable} architecture. 

The 3D VAE encoder $\mathcal{E}(\cdot)$ compresses both spatial and temporal dimension through temporally causal convolutions, mapping an input video $\mathbf{x} \in \mathbb{R}^{(4f+1)\times 8h \times 8w \times c}$ to a latent representation $\mathbf{z}_{vision} = \mathcal{E}(\mathbf{x})\in\mathbb{R}^{(f+1)\times h \times w\times d}$, where $d$ denotes the hidden dimension. 

The transformer processes a concatenated sequence $\mathbf{z} = [\mathbf{z}_{text}, \mathbf{z}_{vision}]$, where text tokens $\mathbf{z}_{text}$ are text embeddings. It utilizes 3D full attention for joint spatio-temporal modeling, incorporating the 3D rotary position embedding (3D-RoPE) during the computation of the query and key. Additionally, the model implements modality-specific adaptive layernorm, where video and text tokens are normalized independently and then modulated by learned scale and shift factors conditioned on their respective modalities. The denoising objective is defined in \cref{denoise_loss}:
\begin{equation}
    \mathcal{L} = \left\|\mathbf{\epsilon} - \mathbf{\epsilon}_\theta(\mathbf{z}_t, t, \mathbf{c})\right\|^2,
    \label{denoise_loss}
\end{equation}
where $\mathbf{\epsilon}$ denotes the noise added to the latent representations,  $t$ represents the diffusion timestep, and $\mathbf{c}$ represents the conditioning signal.

% The VAE compresses spatial and temporal dimensions through temporally causal convolutions. Specifically, the encoder $\mathcal{E}(\cdot)$ maps an input video $\mathbf{x} \in \mathbb{R}^{(4f+1)\times 8h \times 8w \times c}$ to a latent representation $\mathbf{z}_{vision} = \mathcal{E}(\mathbf{x})\in\mathbb{R}^{(f+1)\times h \times w\times d}$, where $d$ denotes the hidden dimension. 

% The expert transformer processes a concatenated sequence $\mathbf{z} = [\mathbf{z}_{text}, \mathbf{z}_{vision}]$, where text tokens $\mathbf{z}_{text}$ are T5-encoded text embeddings. Unlike conventional approaches using separate spatial and temporal attention, it utilizes 3D full attention for joint spatio-temporal modeling, incorporating the 3D rotary position embedding (3D-RoPE) during the computation of the query and key. Additionally, the model implements modality-specific adaptive layernorm,  where video and text tokens are normalized independently and then modulated by learned scale and shift factors conditioned on their respective modalities. The denoising objective is defined in \cref{denoise_loss}:
% \begin{equation}
%     \mathcal{L} = \left\|\mathbf{\epsilon} - \mathbf{\epsilon}_\theta(\mathbf{z}_t, t, \mathbf{c})\right\|^2,
%     \label{denoise_loss}
% \end{equation}
% where $\mathbf{\epsilon}$ denotes the noise added to the latent representations,  $t$ represents the diffusion timestep, and $\mathbf{c}$ represents the conditioning signal.

\subsection{Unified Conditional Embedding Module}
\label{sec3.2}
In various visual generative tasks, inputs are often multi-modal(\eg, text and image), while output is image modality. Therefore, the key to unified visual generation lies in how to effectively fuse multi-modal conditions and avoid confusing conditional and generated images with each other. In RealGeneral, we design the UCE module to align multi-modal conditions, containing two complementary pathways, \ie, the Subject-Specific Embedding layer for inter-modal conditional semantic alignment and the Condition Embedding layer for intra-modal generative semantic distinction.

 As depicted in \cref{framework}, 3D VAE separately encodes the condition and target images, producing latent codes $\mathbf{c}_{cond} \in \mathbb{R}^{h \times w \times d}$ and $\mathbf{x}_{target} \in \mathbb{R}^{h \times w \times d}$, respectively. Then in the UCE module, given a text embedding, we first extract subject-related embeddings $\mathbf{c}_{instance} \in \mathbb{R}^{k}$ using keyword matching (\eg, ``\textit{pot}'' in \cref{framework}). The Subject-Specific Embedding layer then maps these embeddings to the visual latent space, denoted as $\mathbf{c'}_{instance} \in \mathbb{R}^{d}$. This process explicitly anchors textual subject descriptions to corresponding spatial regions in the condition image, thereby aligning textual and visual conditions. Additionally, the Condition Embedding layer applies a learnable task-specific bias $\mathbf{b}_c \in \mathbb{R}^{d}$ to reposition the condition embedding distribution, thereby better aligning the condition image with the target image. The enhanced condition token is obtained by combining the original embedding $\mathbf{c}_{\text{cond}}$ with these adjustments, as \cref{input formula} formulated.
\begin{equation}
    \mathbf{c}_{\text{cond}} = \mathbf{c}_{\text{cond}} + \mathbf{c'}_{instance} + \mathbf{b}_c.
    \label{input formula}
\end{equation}

Before patchify operation, the vision sequence $\mathbf{z}_{vision} \in \mathbb{R}^{2 \times h \times w \times d}$ is formed by concatenating $\mathbf{c}_{cond}$ and $\mathbf{x}_{target}$. The patchify operation in CogVideoX divides the input sequence along spatial and temporal dimensions using patch factors $p=2$. This process compresses the temporal dimension, leading to the intermingling of the latent tokens for $\mathbf{z}_{cond}$ and $\mathbf{z}_{target}$. Such mixing obscures the distinct temporal features inherent to each frame. To mitigate this, we propose a simple replication strategy, \ie, we replicate the latent tokens for both $\mathbf{c}_{cond}$ and $\mathbf{x}_{target}$ along the temporal dimension. Finally, $\mathbf{c}_{cond}$, $\mathbf{x}_{target}$ are concatenated into the input sequence for the Unified Stream DiT Block:
\begin{equation}
    \mathbf{z} = [\mathbf{c}_{text};  \mathbf{c}_{cond}; \mathbf{x}_{target}].
\end{equation}

\subsection{Separated Condition AdaLN Module}
\label{sec3.3}
We introduce the SC-AdaLN module, a novel mechanism to address the inherent conflict between multi-modal conditions in image generation tasks. Existing video generation frameworks employ two AdaLN modules to handle text and video inputs separately. This leads to interference between the condition frame and target frame due to their shared video AdaLN parameters. Our SC-AdaLN module resolves this by decoupling this process into three independent branches, explicitly disentangling three distinct modalities:

\begin{enumerate}
    \item \textbf{Text AdaLN} leverages the semantic information from text embeddings to provide global semantic guidance for the generation process, ensuring text-image alignment. This branch remains unchanged from the original.
    \item \textbf{Condition Frame AdaLN} inherits the parameters from the original video AdaLN to preserve spatial-temporal modeling capabilities. It specializes in extracting style, structure, and local details from the condition image.
    \item \textbf{Target frame AdaLN} also inherits the parameters from the original video AdaLN, which learns transition patterns between the condition and target frames. 
\end{enumerate}

To modulate each branch according to the current timestep $t$, we compute adaptive scale and shift parameters and residual gate via linear layers. Specifically, for each branch $\mathbf{k}\in\{\mathbf{c}_{text}, \mathbf{c}_{cond}, \mathbf{x}_{target}\}$, we have:
\begin{equation}
    \left[\gamma_\mathbf{k},\ \beta_\mathbf{k}, g_\mathbf{k}\right] = f_\mathbf{k}(t),
    \label{mlp}
\end{equation}
where $f_\mathbf{k}$ denotes a MLP that outputs the modulation coefficients, scale $\gamma_\mathbf{k}\in \mathbb{R}^d$, shift $\beta_\mathbf{k}\in \mathbb{R}^d$, and a residual gate $g_\mathbf{k}\in \mathbb{R}^d$ conditioned on $t$.

Then for each branch, the normalized feature is calculated as:
\begin{equation}
    \operatorname{AdaLN}_\mathbf{k}(\mathbf{k}; t) = \operatorname{LN}(\mathbf{k}) \odot (1 + \gamma_\mathbf{k}) + \beta_\mathbf{k},
    \label{adaln}
\end{equation}
where $\operatorname{LN}(\cdot)$ denotes standard layer normalization and $\odot$ represents element-wise multiplication. After applying AdaLN to each branch, the outputs are concatenated into a unified sequence.

\begin{figure}
      \centering
      \includegraphics[width=1\linewidth]{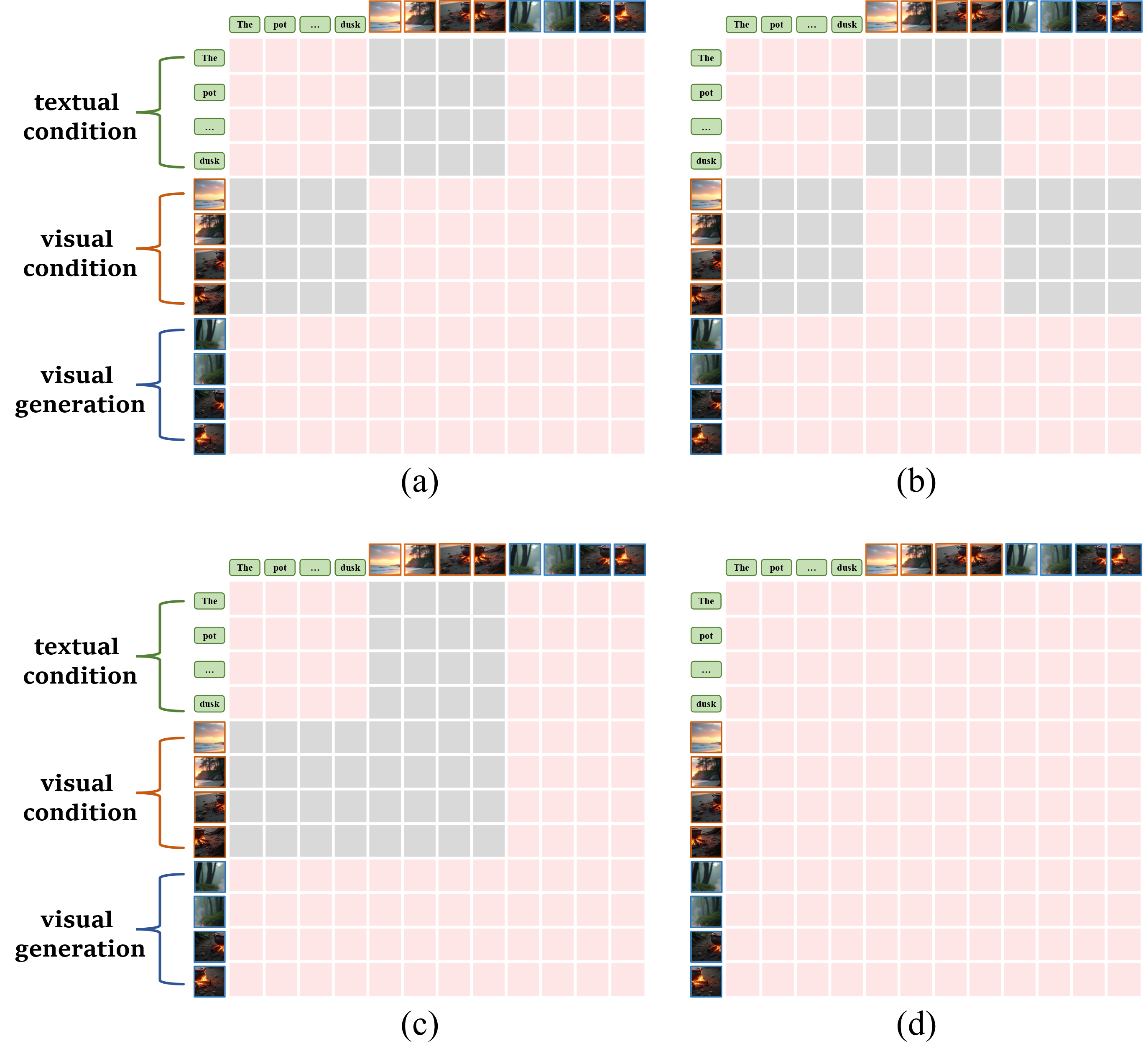}
    \caption{Various attention mask strategies. Red for interaction, and gray for blocked. (a) constrains the interaction between text and condition image. Based on (a), (b) further limits the cross-attention from the target image to the condition image, and (c) restricts the self-attention of the condition image. (d) is the original configuration without any mask.}
    \label{fig:mask}
\end{figure}

% \begin{figure}
%     \centering 
%   \begin{subfigure}{0.23\linewidth}
%   \includegraphics[width=1\linewidth]{image/attention_mask_1.png}
%   \caption{}
%   \label{fig:mask_a}
%   \end{subfigure}
%   \hfill
%   \begin{subfigure}{0.23\linewidth}
%   \includegraphics[width=1\linewidth]{image/attention_mask_2.png}
%   \caption{}
%   \end{subfigure}
%   \begin{subfigure}{0.23\linewidth}
%   \includegraphics[width=1\linewidth]{image/attention_mask_3.png}
%   \caption{}
%   \end{subfigure}
%   \hfill
%   \begin{subfigure}{0.23\linewidth}
%   \includegraphics[width=1\linewidth]{image/attention_mask_4.png}
%   \caption{}
%   \end{subfigure}
%   \caption{Various attention mask strategies. (a) constrains the interaction between text and condition image. Builds on (a), (b) further limits the cross-attention from the target image to the condition image, and (c) restricts the self-attention of the condition image. (d) is the original configuration without any mask.}
%   \label{fig:mask}
% \end{figure}

\subsection{Frame-Condition Decoupling Module}
\label{sec3.4}
% We propose the FCD module to mitigate the multi-modal condition confusion. The module consists of two key components: the Frame-Condition Decoupling Attention (FCD Attention), which addresses the cross-modal interference, and the Unified Image Rotary Embedding, which enhances the scalability of the approach.

We propose the FCD module to mitigate multi-modal condition confusion, mainly containing the Frame-Condition Decoupling Attention (FCD Attention), which addresses the cross-modal condition interference.

In general-purpose image generation tasks, textual input primarily conveys the semantic details of the target frame, while the condition frame provides structural or stylistic guidance. In CogVideoX, text and visual tokens are jointly processed using 3D full attention, allowing unrestricted cross-modal interactions. However, this direct interaction between text tokens and condition tokens leads to the blending of semantic and structural information, negatively affecting the quality of the generated frame.

To address this issue, we introduce the FCD Attention mechanism,  which explicitly restricts interactions between text and condition tokens. Specifically, we employ a custom attention mask, $M$, to control these interactions precisely. The mask assigns an extremely negative value to the similarity scores between text tokens and condition tokens during attention computation, effectively decoupling the multi-modal condition interactions. This ensures that textual input influences only the target frame, while the condition frame preserves its structural properties without being overwhelmed by semantic details.

To validate the effectiveness of our design, we test various masking strategies, as illustrated in \cref{fig:mask}. Among these, the configuration shown in \textit{(a)} achieves the best performance. The mask is defined as:
\begin{equation}
    M_{ij}= \left\{
    \begin{aligned}
    -\infty, &\forall\left(i,j\right)\in\left(\mathbf{c}_{text}\times\mathbf{c}_{cond}\right)\cup\left(\mathbf{c}_{cond}\times\mathbf{c}_{text}\right). \\
    0, & otherwise.
    \end{aligned}
    \right.
    \label{mask}
\end{equation}

Formally, the modified attention operation is defined as:
\begin{equation}
    \operatorname{Attention}(Q, K, V)=\operatorname{Softmax}\left(\frac{Q K^{T}}{\sqrt{d}}+M\right) V.
\end{equation}

%\textbf{Unified Image Rotary Embedding.} To improve the scalability of our method, we introduce a unified rotary embedding for both condition and target frames. Although the training paradigm uses a single condition image, this embedding approach aligns the rotary embeddings across frames, enabling the model to process multiple condition images during inference without enforcing a strict temporal order.

\subsection{Task-Specific LoRA}
\label{sec3.5}
To efficiently adapt the billion-parameter CogVideoX, we employ LoRA~\cite{hu2021lora} for fine-tuning. Specifically, for each task, we inject LoRA parameters exclusively into our proposed Separated Condition AdaLN Module and Frame-Condition Decoupling Module.
\section{Experiments}

\subsection{Implementation Details} 
\textbf{Task.} In our experimental evaluation, we selected three tasks to evaluate our framework: subject-driven text-to-image generation, canny-to-image, and depth-to-image. The subject-driven generation is a challenging task, explored in many works~\cite{gal2022image, ruiz2023dreambooth, kumari2023multi, chen2024customcontrast}, evaluating the model’s ability to generate images related to a specific subject based on conditional inputs. The canny-to-image and depth-to-image tasks assess spatial understanding through edge and depth conditioning, respectively, thereby evaluating structural fidelity and three-dimensional awareness. These tasks demonstrate the model’s versatility and precision in processing diverse input modalities and generating corresponding outputs, highlighting its generalizability.

\textbf{Dataset.}
For subject-driven text-to-image generation, we use FLUX to generate paired images of the same object with variations in scenes and poses, then filter Subject200k~\cite{tan2024ominicontrol} to obtain 40K high-quality pairs, yielding a 260K two-frame video dataset. For spatially-aligned tasks (\eg, canny-to-image, depth-to-image), we extract a subset from 512-2M~\cite{tan2024ominicontrol}, forming 310K videos, fewer than those used in previous models. All videos are resized to $512\times512$. Details refer to supplementary material.

% For subject-driven text-to-image generation, we construct a dataset of paired $1024\times1024$ images depicting the same subject with variations in the scene, pose, and other attributes using FLUX. These pairs are assembled into 220,000 videos. Additionally, we evaluate the quality of the image pairs from the Subject200k~\cite{tan2024ominicontrol} dataset using ChatGPT-4o and retain 40,000 high-quality pairs. Finally, we use 260,000 two-frame videos exhibiting high subject and text consistency for subject-driven generation. For spatially-aligned tasks (\ie, canny-to-image, depth-to-image), we utilize a subset of the 512-2M~\cite{tan2024ominicontrol} dataset, which comprises approximately 310,000 images. We then generate corresponding conditional images, ultimately forming 310,000 videos. All videos are resized to $512\times512$.

\textbf{Benchmark.} For customized generation, we follow previous work~\cite{wei2023elite, chen2024customcontrast}, using 20 testing images in DreamBench~\cite{ruiz2023dreambooth} and 20 prompts. For spatially-aligned tasks, we use the COCO2017 validation set (5000 images) resized to $512\times 512$, with descriptions generated by ChatGPT-4o. 

\textbf{Metric.} For subject-driven generation, we use CLIP-I, DINO~\cite{oquab2023dinov2} to calculate subject similarity and CLIP-T to measure semantic consistency. To mitigate background interference, we compute the subject similarity after segmenting the reference and generated subjects using GroundedSAM~\cite{ren2024grounded}. For spatially-aligned tasks, we use FID~\cite{heusel2017gans} and SSIM to measure image quality.

\textbf{Training and Inference.} Our model, based on CogVideoX1.5 and fine-tuned using LoRA with a rank and alpha of 256, comprises 892 million parameters. Training is performed over 4000 iterations across 2 days on 2 A800 GPUs, with a batch size of 44. We utilize an AdamW optimizer with a learning rate of 0.0001 and betas set to 0.9 and 0.95. During inference, we apply classifier-free guidance for the text condition, with $\omega$ values set to 6 for customization and 2 for other tasks. The denoising step is set at 50.

% Our model is built upon CogVideoX1.5, using LoRA to fine-tune. The LoRA rank and alpha are set to 256, with a total of 892M parameters. The model is trained on 2 A800 GPUs for 4000 iterations over 2 days. We employ a batch size of 44 and use an AdamW optimizer with a learning rate of 0.0001 and betas set to 0.9 and 0.95. During inference, we use classifier-free guidance with respect to text condition, $\omega$ set to 6 for customization and 2 for other tasks. The denoising step is 50. 

\subsection{Main Results}
\begin{figure*}[t]
    \centering
    \includegraphics[width=1\linewidth]{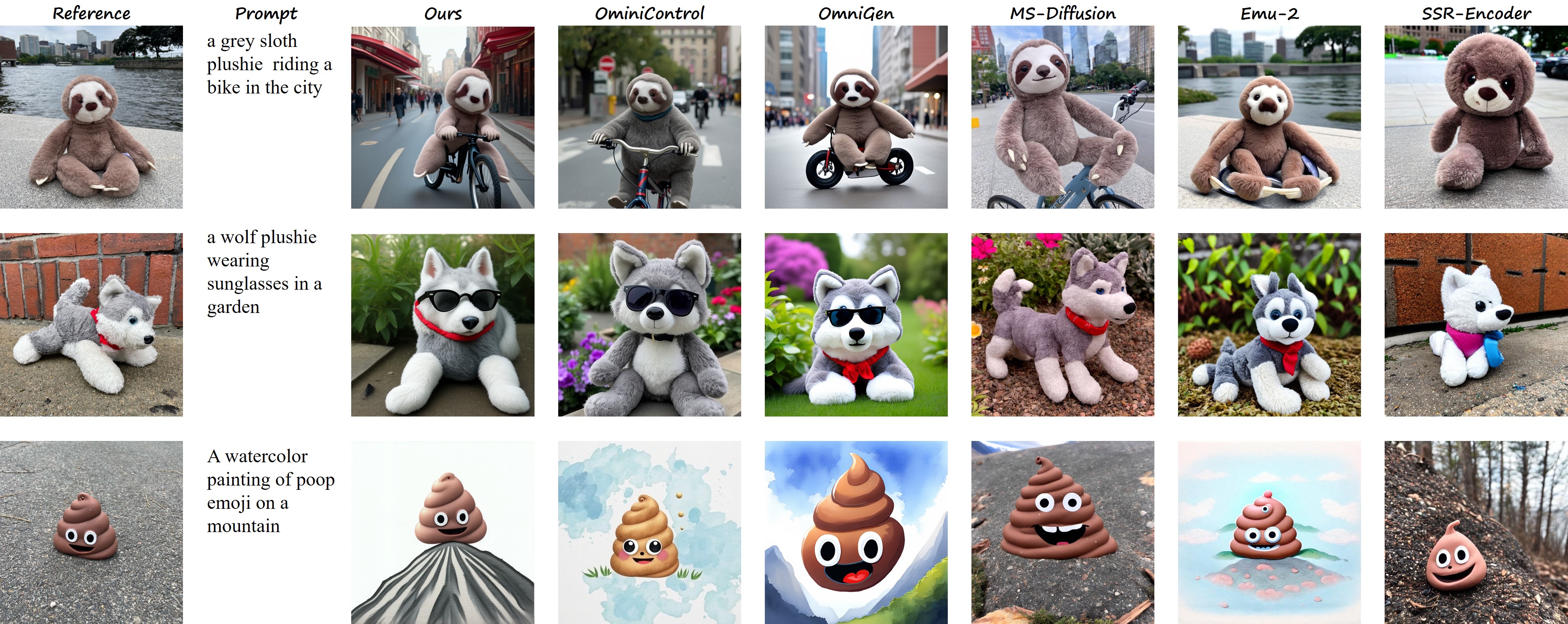}
    \caption{Qualitative comparison with existing methods, consisting of unified image generation models and customization-specific models. The results demonstrate that RealGeneral exhibited superior performance in terms of subject consistency and text consistency.}
    \label{ip-case}
\end{figure*}

\textbf{Customized Image generation.} In \cref{ip-case}, we present qualitative comparisons with existing models, including both unified image generation models (OmniControl, OmniGen) and customization-specific models (MS-Diffusion, Emu2, SSR-Encoder). Compared to other methods, RealGeneral produces more precise details and better adherence to the prompt, underlining its effectiveness in both subject consistency and text controllability.

The quantitative results are shown in \cref{IP-table}. Compared to existing methods, our method demonstrates significant improvement in image similarity. The image similarity is notably higher, with our approach achieving a CLIP-I score of 0.849 and a DINO score of 0.668, surpassing the second-best results by 0.054 and 0.085, respectively. At the same time, our method attains a CLIP-T score of 0.314, outperforming other methods.

\begin{table}
  \centering
  \small
  \resizebox{\linewidth}{!}{ 
      \begin{tabular}{lcccc}
        \toprule
        Method & Base-Model & CLIP-I$\uparrow$ & DINO$\uparrow$ & CLIP-T$\uparrow$ \\
        \midrule
        BLIP-Diffusion~\cite{li2023blip} & SD 1.5 & 0.768 & 0.535 & 0.278  \\
        ELITE~\cite{wei2023elite} & SD 1.5 & 0.762 & 0.533 & 0.291\\
        SSR-Encoder~\cite{zhang2024ssr} & SD 1.5 & 0.767 & 0.524 & 0.303\\
        IP-Adapter~\cite{ye2023ip} & SDXL & 0.790 & 0.570 & 0.289 \\
        EMU2~\cite{sun2024generative} & SDXL & \colorbox{palePurple}{0.795} & \colorbox{palePurple}{0.583} & 0.298 \\
        MS-Diffusion~\cite{wang2025msdiffusion} & SDXL & 0.772 & 0.540 & \colorbox{palePurple}{0.312} \\
        Ours & CogVideoX & \colorbox{paleRed}{0.849} & \colorbox{paleRed}{0.668} & \colorbox{paleRed}{0.314}\\
        \bottomrule
      \end{tabular}
    }
  \caption{Quantitative results with existing methods on customization task. We highlight the \colorbox{paleRed}{best} and \colorbox{palePurple}{second-best} values for each metric.}
  \label{IP-table}
\end{table}

\begin{figure}
    \centering
    \includegraphics[width=1\linewidth]{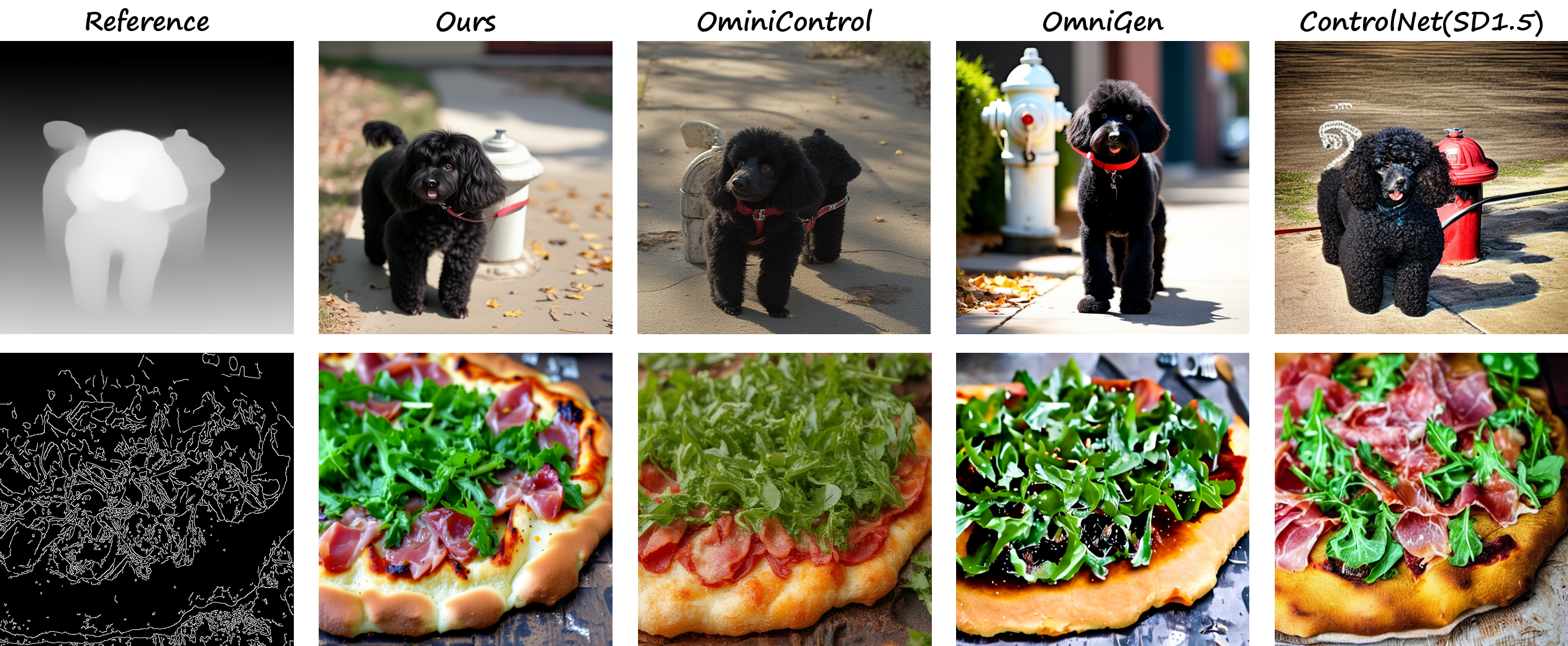}
    \caption{Qualitative comparison with existing methods on canny-to-image task. Our method is more consistent with the given depth or canny map than other methods.}
    \label{compare_canny}
\end{figure}

\textbf{Spatially-aligned tasks.} In \cref{compare_canny}, we compare with existing methods. RealGeneral exhibits enhanced consistency with the provided canny edge or depth maps, generating more detailed and controllable images. Notably, because RealGeneral is based on a large-scale pre-trained video model, it essentially acquires more general visual knowledge and has a better understanding of visual relationships than those methods using image base model, as shown in \cref{compare_canny} row 1 (\ie, dog case). 

The quantitative results are presented in \cref{spatial}. For the canny-to-image task, RealGeneral attains superior FID and SSIM scores compared to existing models. For the depth-to-image task, RealGeneral achieves a balanced performance in terms of FID and SSIM, attaining the second-best for both metrics. In addition, because the depth and canny maps contain spatial information, the CLIP-T of the generated images is very close (both 0.328). We don't show this metric in the table.

% \begin{figure}
%     \centering
%     \includegraphics[width=1\linewidth]{ICCV2025-Author-Kit-Feb/image/compare_depth.png}
%     \caption{Qualitative comparison with existing methods on depth-to-image task.}
%     \label{compare_depth}
% \end{figure}

\begin{table}
  \centering
  \small
  \resizebox{\linewidth}{!}{ 
      \begin{tabular}{clccc}
        \toprule
        Condition & Method & Base-Model & FID$\downarrow$ & SSIM$\uparrow$ \\
        \midrule
        \multirow{5}{*}{Canny} & ControlNet~\cite{zhang2023adding} & SD1.5 & \colorbox{palePurple}{18.79} & 0.28 \\
        & ControlNet~\cite{zhang2023adding} & FLUX & 98.68 & 0.25  \\
        & OmniGen~\cite{xiao2024omnigen} & - & 22.51 & 0.34 \\
        & OminiControl~\cite{tan2024ominicontrol} & FLUX & 20.63 & \colorbox{palePurple}{0.40} \\
        & Ours & CogVideoX & \colorbox{paleRed}{17.50} & \colorbox{paleRed}{0.44} \\
        \midrule
        \multirow{5}{*}{Depth} & ControlNet~\cite{zhang2023adding} & SD1.5 & 25.12 & 0.24 \\
        & ControlNet~\cite{zhang2023adding} & FLUX & 62.20 & 0.26 \\
        & OmniGen~\cite{xiao2024omnigen}  & - & \colorbox{paleRed}{21.62} & 0.25 \\
        & OminiControl~\cite{tan2024ominicontrol} & FLUX & 27.26 & \colorbox{paleRed}{0.39} \\
        & Ours & CogVideoX & \colorbox{palePurple}{23.40} & \colorbox{palePurple}{0.35}  \\
        \bottomrule
      \end{tabular}
      }
  \caption{Quantitative results with existing method on canny-to-image, depth-to-image generation tasks.}
  \label{spatial}
\end{table}

\subsection{Base Model Performance}
\begin{figure}
    \centering
    \includegraphics[width=1\linewidth]{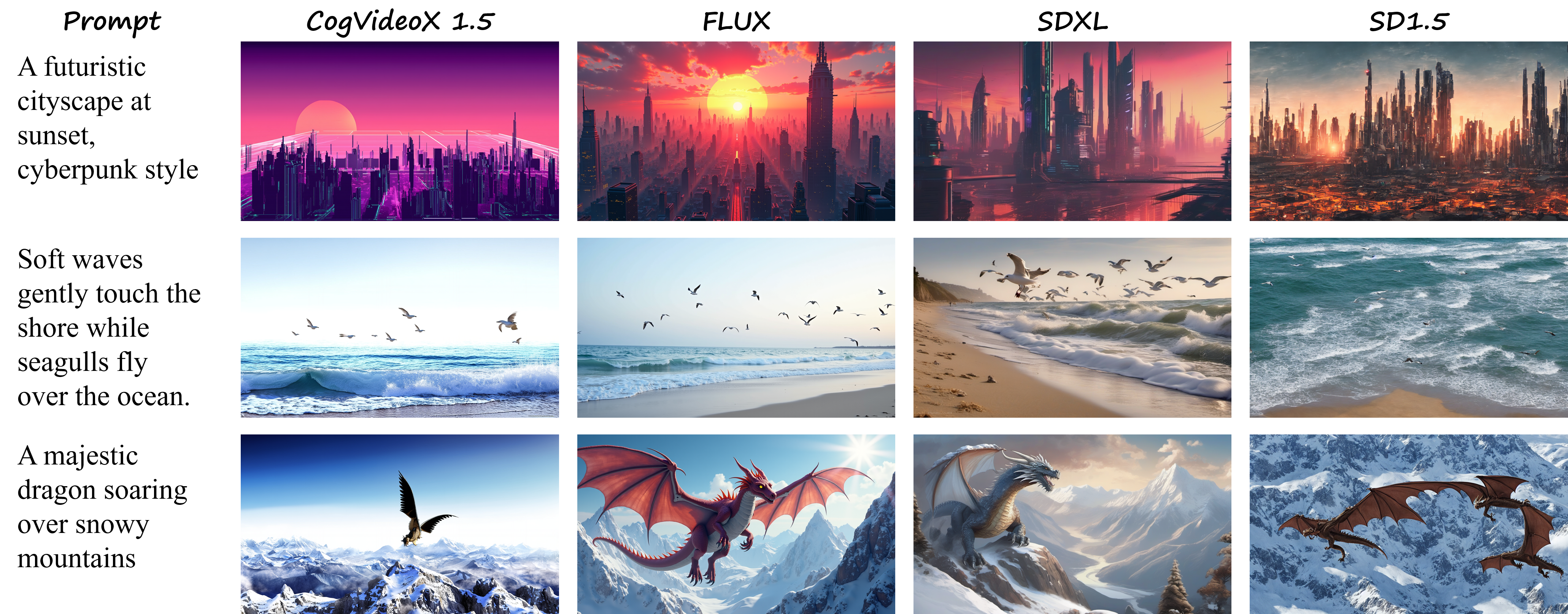}
    \caption{Comparison of image generation results between CogVideoX1.5 and other image generation models.}
    \label{base-model-case}
\end{figure}

We compare the image generation performance between CogVideoX1.5 and image base models (SD1.5, SDXL, FLUX), as shown in \cref{base-model-case}. The results present that CogVideoX1.5 achieves comparable image quality with SD1.5, but significantly weaker than SDXL and FLUX. Nevertheless, RealGeneral still attains better or comparable performance compared to existing methods based on pre-trained image models, demonstrating the effectiveness of the pre-training video model and the viability of unifying visual generation through task-agnostic videos.

\subsection{Ablation Studies}
\textbf{Effect of each module.} To validate the effectiveness of our proposed modules, we conduct the ablation experiments on the Unified Conditional Embedding(UCE), Separated Condition AdaLN(SC-AdaLN), and Frame-Condition Decoupling(FCD) modules, with the results shown in \cref{module-ablation}. Removing UCE module results in a significant decrease in CLIP-I (0.849 $\to$ 0.833), indicating its critical role in aligning multi-modal semantics. Similarly, the removal of SC-AdaLN or FCD module leads to reductions in CLIP-I by 0.019 and 0.022, respectively, showcasing the importance of decoupling multi-modal conditions. Although these modifications cause a slight decrease in CLIP-T, the improvement in topic similarity remains noteworthy.

% \begin{figure}
%     \centering
%     \includegraphics[width=1\linewidth]{image/compare_ablation.png}
%     \caption{Ablation study for our proposed module. The full version presents more subject consistency while demonstrating comparable textual controllability.}
%     \label{ablation-case}
% \end{figure}

\begin{table}
  \centering
      \begin{tabular}{lccc}
        \toprule
        Setting & CLIP-I$\uparrow$ & DINO$\uparrow$ & CLIP-T$\uparrow$  \\
        \midrule
        w/o UCE & 0.833 & 0.666 & \textbf{0.320} \\
        w/o SC-AdaLN & 0.830 & 0.666 & 0.316 \\
        w/o FCD & 0.827 & 0.665 & 0.318 \\
        \rowcolor{lightblue}
        full version & \textbf{0.849} & \textbf{0.668} & 0.314 \\
        \bottomrule
      \end{tabular}
  \caption{Effect of \textit{Unified Conditional Embedding(UCE), Separated Condition AdaLN(SC-AdaLN), Frame-Condition Decoupling(FCD)} modules. Removing any of the above modules causes a significant reduction in subject consistency while attaining a subtle increase in textual controllability.}
  \label{module-ablation}
\end{table}

\textbf{Impact of various attention mask strategies.} To assess the impact of different attention mask strategies on model performance, we conduct ablation experiments as shown in \cref{mask-ablation}. Among the several attention mask strategies depicted in \cref{fig:mask}, Mask A, which blocks the interaction between the textual input and the condition image, achieves the highest scores in subject similarity with only a minimal reduction in textual similarity. In contrast, both Mask B and the no-mask result in lower subject consistency, and Mask C significantly impairs the model's performance in both subject consistency and textual controllability by restricting the condition image's self-attention. The qualitative results are shown in supplementary material.

% \begin{figure}
%     \centering
%     \includegraphics[width=1\linewidth]{image/compare_mask.png}
%     \caption{Qualitative results of various attention masks. The mask is depicted in \cref{fig:mask}.}
%     \label{ablation-mask}
% \end{figure}

\begin{table}
  \centering
    \begin{tabular}{lccc}
        \toprule
        Setting & CLIP-I$\uparrow$ & DINO$\uparrow$ & CLIP-T$\uparrow$  \\
        \midrule
        Mask B & 0.827 & 0.662 & \textbf{0.321} \\
        Mask C & 0.779 & 0.650 &  0.312\\
        no mask & 0.827 & 0.665 & 0.318 \\
        \rowcolor{lightblue}
        Mask A & \textbf{0.849} & \textbf{0.668} & 0.314 \\
        \bottomrule
      \end{tabular}
  \caption{Ablation study on various attention masks.}
  \label{mask-ablation}
\end{table}

\section{Conclusion \& Future Work}
In this paper, we propose RealGeneral, a novel framework that bridges the gap between video generation models and unified visual synthesis tasks. We unlock the inherent potential of pre-trained video models for diverse image generation tasks by reformulating conditional image generation as temporal in-context learning through temporal modeling. This is achieved through three core modules: (1) UCE module that aligns multi-modal semantics, (2) SC-AdaLN that disentangles feature modulation across text, conditional frames, and target frames, (3) FCD module that prevents conditional semantic interference. Together, these components form a solution for unified image generation tasks.

\textbf{Future Work.} Future research will explore several directions to further advance RealGeneral. First, we plan to validate the framework on video models with higher generation quality and larger pre-training to assess its generalizability. Second, we will focus on developing enhanced video foundation models that generate higher-quality image synthesis. Finally, we aim to develop a fully integrated, universal framework instead of task-specific LoRA fine-tuning.

% Although RealGeneral demonstrates promising potential, it still has several limitations. First, the framework's effectiveness has been validated only on CogVideoX, and its generalizability to other video models remains unverified. Second, the image quality is suboptimal compared to those models based on FLUX, which remains enhancement space. Finally, while RealGeneral introduces the concept of unified visual generation, its current implementation relies on task-specific LoRA fine-tuning rather than a fully integrated, universal framework.
\section*{Acknowledgment}
This research is supported by Artificial Intelligence National Science and Technology Major Project 2023ZD0121200, and National Natural Science Foundation of China under Grant 62222212 and 623B2094.
{
    \small
    \bibliographystyle{ieeenat_fullname}
    \bibliography{main}

\begin{thebibliography}{59}
\providecommand{\natexlab}[1]{#1}
\providecommand{\url}[1]{\texttt{#1}}
\expandafter\ifx\csname urlstyle\endcsname\relax
  \providecommand{\doi}[1]{doi: #1}\else
  \providecommand{\doi}{doi: \begingroup \urlstyle{rm}\Url}\fi

\bibitem[Achiam et~al.(2023)Achiam, Adler, Agarwal, Ahmad, Akkaya, Aleman, Almeida, Altenschmidt, Altman, Anadkat, et~al.]{achiam2023gpt}
Josh Achiam, Steven Adler, Sandhini Agarwal, Lama Ahmad, Ilge Akkaya, Florencia~Leoni Aleman, Diogo Almeida, Janko Altenschmidt, Sam Altman, Shyamal Anadkat, et~al.
\newblock Gpt-4 technical report.
\newblock \emph{arXiv preprint arXiv:2303.08774}, 2023.

\bibitem[Bai et~al.(2024)Bai, Geng, Mangalam, Bar, Yuille, Darrell, Malik, and Efros]{bai2024sequential}
Yutong Bai, Xinyang Geng, Karttikeya Mangalam, Amir Bar, Alan~L Yuille, Trevor Darrell, Jitendra Malik, and Alexei~A Efros.
\newblock Sequential modeling enables scalable learning for large vision models.
\newblock In \emph{Proceedings of the IEEE/CVF Conference on Computer Vision and Pattern Recognition}, pages 22861--22872, 2024.

\bibitem[Blattmann et~al.(2023)Blattmann, Rombach, Ling, Dockhorn, Kim, Fidler, and Kreis]{blattmann2023align}
Andreas Blattmann, Robin Rombach, Huan Ling, Tim Dockhorn, Seung~Wook Kim, Sanja Fidler, and Karsten Kreis.
\newblock Align your latents: High-resolution video synthesis with latent diffusion models.
\newblock In \emph{Proceedings of the IEEE/CVF conference on computer vision and pattern recognition}, pages 22563--22575, 2023.

\bibitem[Brown et~al.(2020)Brown, Mann, Ryder, Subbiah, Kaplan, Dhariwal, Neelakantan, Shyam, Sastry, Askell, et~al.]{brown2020language}
Tom Brown, Benjamin Mann, Nick Ryder, Melanie Subbiah, Jared~D Kaplan, Prafulla Dhariwal, Arvind Neelakantan, Pranav Shyam, Girish Sastry, Amanda Askell, et~al.
\newblock Language models are few-shot learners.
\newblock \emph{Advances in neural information processing systems}, 33:\penalty0 1877--1901, 2020.

\bibitem[Chang et~al.(2023)Chang, Zhang, Barber, Maschinot, Lezama, Jiang, Yang, Murphy, Freeman, Rubinstein, et~al.]{chang2023muse}
Huiwen Chang, Han Zhang, Jarred Barber, AJ Maschinot, Jose Lezama, Lu Jiang, Ming-Hsuan Yang, Kevin Murphy, William~T Freeman, Michael Rubinstein, et~al.
\newblock Muse: Text-to-image generation via masked generative transformers.
\newblock \emph{arXiv preprint arXiv:2301.00704}, 2023.

\bibitem[Chen et~al.(2024{\natexlab{a}})Chen, Wang, Yang, and Lim]{chen2024omnicreator}
Haodong Chen, Lan Wang, Harry Yang, and Ser-Nam Lim.
\newblock Omnicreator: Self-supervised unified generation with universal editing.
\newblock \emph{arXiv preprint arXiv:2412.02114}, 2024{\natexlab{a}}.

\bibitem[Chen et~al.(2024{\natexlab{b}})Chen, Huang, Chen, Zheng, Zhang, and Mao]{chen2024customcontrast}
Nan Chen, Mengqi Huang, Zhuowei Chen, Yang Zheng, Lei Zhang, and Zhendong Mao.
\newblock Customcontrast: A multilevel contrastive perspective for subject-driven text-to-image customization.
\newblock \emph{arXiv preprint arXiv:2409.05606}, 2024{\natexlab{b}}.

\bibitem[Chen et~al.(2024{\natexlab{c}})Chen, Zhang, Zhang, Zhou, Kim, Liu, Li, Zhang, Zhao, Wang, et~al.]{chen2024unireal}
Xi Chen, Zhifei Zhang, He Zhang, Yuqian Zhou, Soo~Ye Kim, Qing Liu, Yijun Li, Jianming Zhang, Nanxuan Zhao, Yilin Wang, et~al.
\newblock Unireal: Universal image generation and editing via learning real-world dynamics.
\newblock \emph{arXiv preprint arXiv:2412.07774}, 2024{\natexlab{c}}.

\bibitem[Dhariwal and Nichol(2021)]{dhariwal2021diffusion}
Prafulla Dhariwal and Alexander Nichol.
\newblock Diffusion models beat gans on image synthesis.
\newblock \emph{Advances in neural information processing systems}, 34:\penalty0 8780--8794, 2021.

\bibitem[Esser et~al.(2021)Esser, Rombach, and Ommer]{esser2021taming}
Patrick Esser, Robin Rombach, and Bjorn Ommer.
\newblock Taming transformers for high-resolution image synthesis.
\newblock In \emph{Proceedings of the IEEE/CVF conference on computer vision and pattern recognition}, pages 12873--12883, 2021.

\bibitem[Gal et~al.(2022)Gal, Alaluf, Atzmon, Patashnik, Bermano, Chechik, and Cohen-Or]{gal2022image}
Rinon Gal, Yuval Alaluf, Yuval Atzmon, Or Patashnik, Amit~H Bermano, Gal Chechik, and Daniel Cohen-Or.
\newblock An image is worth one word: Personalizing text-to-image generation using textual inversion.
\newblock \emph{arXiv preprint arXiv:2208.01618}, 2022.

\bibitem[Guo et~al.(2023)Guo, Yang, Rao, Liang, Wang, Qiao, Agrawala, Lin, and Dai]{guo2023animatediff}
Yuwei Guo, Ceyuan Yang, Anyi Rao, Zhengyang Liang, Yaohui Wang, Yu Qiao, Maneesh Agrawala, Dahua Lin, and Bo Dai.
\newblock Animatediff: Animate your personalized text-to-image diffusion models without specific tuning.
\newblock \emph{arXiv preprint arXiv:2307.04725}, 2023.

\bibitem[Han et~al.(2024)Han, Jiang, Pan, Zhang, Mao, Xie, Liu, and Zhou]{han2024ace}
Zhen Han, Zeyinzi Jiang, Yulin Pan, Jingfeng Zhang, Chaojie Mao, Chenwei Xie, Yu Liu, and Jingren Zhou.
\newblock Ace: All-round creator and editor following instructions via diffusion transformer.
\newblock \emph{arXiv preprint arXiv:2410.00086}, 2024.

\bibitem[Heusel et~al.(2017)Heusel, Ramsauer, Unterthiner, Nessler, and Hochreiter]{heusel2017gans}
Martin Heusel, Hubert Ramsauer, Thomas Unterthiner, Bernhard Nessler, and Sepp Hochreiter.
\newblock Gans trained by a two time-scale update rule converge to a local nash equilibrium.
\newblock \emph{Advances in neural information processing systems}, 30, 2017.

\bibitem[Ho et~al.(2020)Ho, Jain, and Abbeel]{ho2020denoising}
Jonathan Ho, Ajay Jain, and Pieter Abbeel.
\newblock Denoising diffusion probabilistic models.
\newblock \emph{Advances in neural information processing systems}, 33:\penalty0 6840--6851, 2020.

\bibitem[Ho et~al.(2022{\natexlab{a}})Ho, Saharia, Chan, Fleet, Norouzi, and Salimans]{ho2022cascaded}
Jonathan Ho, Chitwan Saharia, William Chan, David~J Fleet, Mohammad Norouzi, and Tim Salimans.
\newblock Cascaded diffusion models for high fidelity image generation.
\newblock \emph{Journal of Machine Learning Research}, 23\penalty0 (47):\penalty0 1--33, 2022{\natexlab{a}}.

\bibitem[Ho et~al.(2022{\natexlab{b}})Ho, Salimans, Gritsenko, Chan, Norouzi, and Fleet]{ho2022video}
Jonathan Ho, Tim Salimans, Alexey Gritsenko, William Chan, Mohammad Norouzi, and David~J Fleet.
\newblock Video diffusion models.
\newblock \emph{Advances in Neural Information Processing Systems}, 35:\penalty0 8633--8646, 2022{\natexlab{b}}.

\bibitem[Hong et~al.(2022)Hong, Ding, Zheng, Liu, and Tang]{hong2022cogvideo}
Wenyi Hong, Ming Ding, Wendi Zheng, Xinghan Liu, and Jie Tang.
\newblock Cogvideo: Large-scale pretraining for text-to-video generation via transformers.
\newblock \emph{arXiv preprint arXiv:2205.15868}, 2022.

\bibitem[Hu et~al.(2021)Hu, Shen, Wallis, Allen-Zhu, Li, Wang, Wang, and Chen]{hu2021lora}
Edward~J Hu, Yelong Shen, Phillip Wallis, Zeyuan Allen-Zhu, Yuanzhi Li, Shean Wang, Lu Wang, and Weizhu Chen.
\newblock Lora: Low-rank adaptation of large language models.
\newblock \emph{arXiv preprint arXiv:2106.09685}, 2021.

\bibitem[Huang et~al.(2022)Huang, Mao, Wang, Wang, and Zhang]{huang2022dse}
Mengqi Huang, Zhendong Mao, Penghui Wang, Quan Wang, and Yongdong Zhang.
\newblock Dse-gan: Dynamic semantic evolution generative adversarial network for text-to-image generation.
\newblock In \emph{Proceedings of the 30th ACM International Conference on Multimedia}, pages 4345--4354, 2022.

\bibitem[Huang et~al.(2023)Huang, Mao, Chen, and Zhang]{huang2023towards}
Mengqi Huang, Zhendong Mao, Zhuowei Chen, and Yongdong Zhang.
\newblock Towards accurate image coding: Improved autoregressive image generation with dynamic vector quantization.
\newblock In \emph{Proceedings of the IEEE/CVF Conference on Computer Vision and Pattern Recognition}, pages 22596--22605, 2023.

\bibitem[Huang et~al.(2024)Huang, Mao, Liu, He, and Zhang]{huang2024realcustom}
Mengqi Huang, Zhendong Mao, Mingcong Liu, Qian He, and Yongdong Zhang.
\newblock Realcustom: Narrowing real text word for real-time open-domain text-to-image customization.
\newblock In \emph{Proceedings of the IEEE/CVF Conference on Computer Vision and Pattern Recognition}, pages 7476--7485, 2024.

\bibitem[Kumari et~al.(2023)Kumari, Zhang, Zhang, Shechtman, and Zhu]{kumari2023multi}
Nupur Kumari, Bingliang Zhang, Richard Zhang, Eli Shechtman, and Jun-Yan Zhu.
\newblock Multi-concept customization of text-to-image diffusion.
\newblock In \emph{Proceedings of the IEEE/CVF conference on computer vision and pattern recognition}, pages 1931--1941, 2023.

\bibitem[Labs(2024)]{flux2024}
Black~Forest Labs.
\newblock Flux: Official inference repository for flux.1 models.
\newblock \url{https://github.com/black-forest-labs/flux}, 2024.
\newblock Accessed: 2024-11-12.

\bibitem[Li et~al.(2023)Li, Li, and Hoi]{li2023blip}
Dongxu Li, Junnan Li, and Steven Hoi.
\newblock Blip-diffusion: Pre-trained subject representation for controllable text-to-image generation and editing.
\newblock \emph{Advances in Neural Information Processing Systems}, 36:\penalty0 30146--30166, 2023.

\bibitem[Li et~al.(2022)Li, Lin, Meng, Ermon, Han, and Zhu]{li2022efficient}
Muyang Li, Ji Lin, Chenlin Meng, Stefano Ermon, Song Han, and Jun-Yan Zhu.
\newblock Efficient spatially sparse inference for conditional gans and diffusion models.
\newblock \emph{Advances in neural information processing systems}, 35:\penalty0 28858--28873, 2022.

\bibitem[Lin et~al.(2024)Lin, Wei, Zhang, Zhuo, Zhao, Huang, Xie, Qiao, Gao, and Li]{lin2024pixwizard}
Weifeng Lin, Xinyu Wei, Renrui Zhang, Le Zhuo, Shitian Zhao, Siyuan Huang, Junlin Xie, Yu Qiao, Peng Gao, and Hongsheng Li.
\newblock Pixwizard: Versatile image-to-image visual assistant with open-language instructions.
\newblock \emph{arXiv preprint arXiv:2409.15278}, 2024.

\bibitem[Liu et~al.(2025)Liu, Wang, Li, van~de Weijer, Khan, Yang, Wang, Yang, and Cheng]{liu2025one}
Tao Liu, Kai Wang, Senmao Li, Joost van~de Weijer, Fahad~Shahbaz Khan, Shiqi Yang, Yaxing Wang, Jian Yang, and Ming-Ming Cheng.
\newblock One-prompt-one-story: Free-lunch consistent text-to-image generation using a single prompt.
\newblock \emph{arXiv preprint arXiv:2501.13554}, 2025.

\bibitem[Mao et~al.(2024)Mao, Huang, Ding, Liu, He, and Zhang]{mao2024realcustom++}
Zhendong Mao, Mengqi Huang, Fei Ding, Mingcong Liu, Qian He, and Yongdong Zhang.
\newblock Realcustom++: Representing images as real-word for real-time customization.
\newblock \emph{arXiv preprint arXiv:2408.09744}, 2024.

\bibitem[Mizrahi et~al.(2023)Mizrahi, Bachmann, Kar, Yeo, Gao, Dehghan, and Zamir]{mizrahi20234m}
David Mizrahi, Roman Bachmann, Oguzhan Kar, Teresa Yeo, Mingfei Gao, Afshin Dehghan, and Amir Zamir.
\newblock 4m: Massively multimodal masked modeling.
\newblock \emph{Advances in Neural Information Processing Systems}, 36:\penalty0 58363--58408, 2023.

\bibitem[Mou et~al.(2024)Mou, Wang, Xie, Wu, Zhang, Qi, and Shan]{mou2024t2i}
Chong Mou, Xintao Wang, Liangbin Xie, Yanze Wu, Jian Zhang, Zhongang Qi, and Ying Shan.
\newblock T2i-adapter: Learning adapters to dig out more controllable ability for text-to-image diffusion models.
\newblock In \emph{Proceedings of the AAAI conference on artificial intelligence}, pages 4296--4304, 2024.

\bibitem[OpenAI(2024)]{openai2024sora}
OpenAI.
\newblock Sora.
\newblock \url{https://openai.com/index/sora/}, 2024.

\bibitem[Oquab et~al.(2023)Oquab, Darcet, Moutakanni, Vo, Szafraniec, Khalidov, Fernandez, Haziza, Massa, El-Nouby, et~al.]{oquab2023dinov2}
Maxime Oquab, Timoth{\'e}e Darcet, Th{\'e}o Moutakanni, Huy Vo, Marc Szafraniec, Vasil Khalidov, Pierre Fernandez, Daniel Haziza, Francisco Massa, Alaaeldin El-Nouby, et~al.
\newblock Dinov2: Learning robust visual features without supervision.
\newblock \emph{arXiv preprint arXiv:2304.07193}, 2023.

\bibitem[Peebles and Xie(2023)]{peebles2023scalable}
William Peebles and Saining Xie.
\newblock Scalable diffusion models with transformers.
\newblock In \emph{Proceedings of the IEEE/CVF International Conference on Computer Vision}, pages 4195--4205, 2023.

\bibitem[Podell et~al.(2023)Podell, English, Lacey, Blattmann, Dockhorn, M{\"u}ller, Penna, and Rombach]{podell2023sdxl}
Dustin Podell, Zion English, Kyle Lacey, Andreas Blattmann, Tim Dockhorn, Jonas M{\"u}ller, Joe Penna, and Robin Rombach.
\newblock Sdxl: Improving latent diffusion models for high-resolution image synthesis.
\newblock \emph{arXiv preprint arXiv:2307.01952}, 2023.

\bibitem[Qin et~al.(2023)Qin, Zhang, Yu, Feng, Yang, Zhou, Wang, Niebles, Xiong, Savarese, et~al.]{qin2023unicontrol}
Can Qin, Shu Zhang, Ning Yu, Yihao Feng, Xinyi Yang, Yingbo Zhou, Huan Wang, Juan~Carlos Niebles, Caiming Xiong, Silvio Savarese, et~al.
\newblock Unicontrol: A unified diffusion model for controllable visual generation in the wild.
\newblock \emph{arXiv preprint arXiv:2305.11147}, 2023.

\bibitem[Raffel et~al.(2020)Raffel, Shazeer, Roberts, Lee, Narang, Matena, Zhou, Li, and Liu]{raffel2020exploring}
Colin Raffel, Noam Shazeer, Adam Roberts, Katherine Lee, Sharan Narang, Michael Matena, Yanqi Zhou, Wei Li, and Peter~J Liu.
\newblock Exploring the limits of transfer learning with a unified text-to-text transformer.
\newblock \emph{Journal of machine learning research}, 21\penalty0 (140):\penalty0 1--67, 2020.

\bibitem[Ramesh et~al.(2021)Ramesh, Pavlov, Goh, Gray, Voss, Radford, Chen, and Sutskever]{ramesh2021zero}
Aditya Ramesh, Mikhail Pavlov, Gabriel Goh, Scott Gray, Chelsea Voss, Alec Radford, Mark Chen, and Ilya Sutskever.
\newblock Zero-shot text-to-image generation.
\newblock In \emph{International conference on machine learning}, pages 8821--8831. Pmlr, 2021.

\bibitem[Ren et~al.(2024)Ren, Liu, Zeng, Lin, Li, Cao, Chen, Huang, Chen, Yan, et~al.]{ren2024grounded}
Tianhe Ren, Shilong Liu, Ailing Zeng, Jing Lin, Kunchang Li, He Cao, Jiayu Chen, Xinyu Huang, Yukang Chen, Feng Yan, et~al.
\newblock Grounded sam: Assembling open-world models for diverse visual tasks.
\newblock \emph{arXiv preprint arXiv:2401.14159}, 2024.

\bibitem[Rombach et~al.(2022)Rombach, Blattmann, Lorenz, Esser, and Ommer]{rombach2022high}
Robin Rombach, Andreas Blattmann, Dominik Lorenz, Patrick Esser, and Bj{\"o}rn Ommer.
\newblock High-resolution image synthesis with latent diffusion models.
\newblock In \emph{Proceedings of the IEEE/CVF conference on computer vision and pattern recognition}, pages 10684--10695, 2022.

\bibitem[Ruiz et~al.(2023)Ruiz, Li, Jampani, Pritch, Rubinstein, and Aberman]{ruiz2023dreambooth}
Nataniel Ruiz, Yuanzhen Li, Varun Jampani, Yael Pritch, Michael Rubinstein, and Kfir Aberman.
\newblock Dreambooth: Fine tuning text-to-image diffusion models for subject-driven generation.
\newblock In \emph{Proceedings of the IEEE/CVF conference on computer vision and pattern recognition}, pages 22500--22510, 2023.

\bibitem[Singer et~al.(2022)Singer, Polyak, Hayes, Yin, An, Zhang, Hu, Yang, Ashual, Gafni, et~al.]{singer2022make}
Uriel Singer, Adam Polyak, Thomas Hayes, Xi Yin, Jie An, Songyang Zhang, Qiyuan Hu, Harry Yang, Oron Ashual, Oran Gafni, et~al.
\newblock Make-a-video: Text-to-video generation without text-video data.
\newblock \emph{arXiv preprint arXiv:2209.14792}, 2022.

\bibitem[Sun et~al.(2023)Sun, Yu, Cui, Zhang, Zhang, Wang, Gao, Liu, Huang, and Wang]{sun2023emu}
Quan Sun, Qiying Yu, Yufeng Cui, Fan Zhang, Xiaosong Zhang, Yueze Wang, Hongcheng Gao, Jingjing Liu, Tiejun Huang, and Xinlong Wang.
\newblock Emu: Generative pretraining in multimodality.
\newblock \emph{arXiv preprint arXiv:2307.05222}, 2023.

\bibitem[Sun et~al.(2024{\natexlab{a}})Sun, Cui, Zhang, Zhang, Yu, Wang, Rao, Liu, Huang, and Wang]{sun2024generative}
Quan Sun, Yufeng Cui, Xiaosong Zhang, Fan Zhang, Qiying Yu, Yueze Wang, Yongming Rao, Jingjing Liu, Tiejun Huang, and Xinlong Wang.
\newblock Generative multimodal models are in-context learners.
\newblock In \emph{Proceedings of the IEEE/CVF Conference on Computer Vision and Pattern Recognition}, pages 14398--14409, 2024{\natexlab{a}}.

\bibitem[Sun et~al.(2024{\natexlab{b}})Sun, Chu, Zhang, Wu, Dong, Zang, Xiong, Lin, and Wang]{sun2024x}
Zeyi Sun, Ziyang Chu, Pan Zhang, Tong Wu, Xiaoyi Dong, Yuhang Zang, Yuanjun Xiong, Dahua Lin, and Jiaqi Wang.
\newblock X-prompt: Towards universal in-context image generation in auto-regressive vision language foundation models.
\newblock \emph{arXiv preprint arXiv:2412.01824}, 2024{\natexlab{b}}.

\bibitem[Tan et~al.(2024)Tan, Liu, Yang, Xue, and Wang]{tan2024ominicontrol}
Zhenxiong Tan, Songhua Liu, Xingyi Yang, Qiaochu Xue, and Xinchao Wang.
\newblock Ominicontrol: Minimal and universal control for diffusion transformer.
\newblock \emph{arXiv preprint arXiv:2411.15098}, 2024.

\bibitem[Villegas et~al.(2022)Villegas, Babaeizadeh, Kindermans, Moraldo, Zhang, Saffar, Castro, Kunze, and Erhan]{villegas2022phenaki}
Ruben Villegas, Mohammad Babaeizadeh, Pieter-Jan Kindermans, Hernan Moraldo, Han Zhang, Mohammad~Taghi Saffar, Santiago Castro, Julius Kunze, and Dumitru Erhan.
\newblock Phenaki: Variable length video generation from open domain textual description.
\newblock \emph{arXiv preprint arXiv:2210.02399}, 2022.

\bibitem[Wang et~al.(2023)Wang, Wang, Cao, Shen, and Huang]{wang2023images}
Xinlong Wang, Wen Wang, Yue Cao, Chunhua Shen, and Tiejun Huang.
\newblock Images speak in images: A generalist painter for in-context visual learning.
\newblock In \emph{Proceedings of the IEEE/CVF Conference on Computer Vision and Pattern Recognition}, pages 6830--6839, 2023.

\bibitem[Wang et~al.(2025)Wang, Fu, Huang, He, and Jiang]{wang2025msdiffusion}
Xierui Wang, Siming Fu, Qihan Huang, Wanggui He, and Hao Jiang.
\newblock {MS}-diffusion: Multi-subject zero-shot image personalization with layout guidance.
\newblock In \emph{The Thirteenth International Conference on Learning Representations}, 2025.

\bibitem[Wang et~al.(2024)Wang, Xia, Chen, Yu, Wang, Gong, and Liu]{wang2024lavin}
Zhaoqing Wang, Xiaobo Xia, Runnan Chen, Dongdong Yu, Changhu Wang, Mingming Gong, and Tongliang Liu.
\newblock Lavin-dit: Large vision diffusion transformer.
\newblock \emph{arXiv preprint arXiv:2411.11505}, 2024.

\bibitem[Wei et~al.(2022)Wei, Wang, Schuurmans, Bosma, Xia, Chi, Le, Zhou, et~al.]{wei2022chain}
Jason Wei, Xuezhi Wang, Dale Schuurmans, Maarten Bosma, Fei Xia, Ed Chi, Quoc~V Le, Denny Zhou, et~al.
\newblock Chain-of-thought prompting elicits reasoning in large language models.
\newblock \emph{Advances in neural information processing systems}, 35:\penalty0 24824--24837, 2022.

\bibitem[Wei et~al.(2023)Wei, Zhang, Ji, Bai, Zhang, and Zuo]{wei2023elite}
Yuxiang Wei, Yabo Zhang, Zhilong Ji, Jinfeng Bai, Lei Zhang, and Wangmeng Zuo.
\newblock Elite: Encoding visual concepts into textual embeddings for customized text-to-image generation.
\newblock In \emph{Proceedings of the IEEE/CVF International Conference on Computer Vision}, pages 15943--15953, 2023.

\bibitem[Xiao et~al.(2024)Xiao, Wang, Zhou, Yuan, Xing, Yan, Wang, Huang, and Liu]{xiao2024omnigen}
Shitao Xiao, Yueze Wang, Junjie Zhou, Huaying Yuan, Xingrun Xing, Ruiran Yan, Shuting Wang, Tiejun Huang, and Zheng Liu.
\newblock Omnigen: Unified image generation.
\newblock \emph{arXiv preprint arXiv:2409.11340}, 2024.

\bibitem[Xu et~al.(2024)Xu, Guo, Wang, Huang, Essa, and Shi]{xu2024prompt}
Xingqian Xu, Jiayi Guo, Zhangyang Wang, Gao Huang, Irfan Essa, and Humphrey Shi.
\newblock Prompt-free diffusion: Taking" text" out of text-to-image diffusion models.
\newblock In \emph{Proceedings of the IEEE/CVF Conference on Computer Vision and Pattern Recognition}, pages 8682--8692, 2024.

\bibitem[Yang et~al.(2024)Yang, Teng, Zheng, Ding, Huang, Xu, Yang, Hong, Zhang, Feng, et~al.]{yang2024cogvideox}
Zhuoyi Yang, Jiayan Teng, Wendi Zheng, Ming Ding, Shiyu Huang, Jiazheng Xu, Yuanming Yang, Wenyi Hong, Xiaohan Zhang, Guanyu Feng, et~al.
\newblock Cogvideox: Text-to-video diffusion models with an expert transformer.
\newblock \emph{arXiv preprint arXiv:2408.06072}, 2024.

\bibitem[Ye et~al.(2023)Ye, Zhang, Liu, Han, and Yang]{ye2023ip}
Hu Ye, Jun Zhang, Sibo Liu, Xiao Han, and Wei Yang.
\newblock Ip-adapter: Text compatible image prompt adapter for text-to-image diffusion models.
\newblock \emph{arXiv preprint arXiv:2308.06721}, 2023.

\bibitem[Yu et~al.(2024)Yu, Chow, Yue, Pan, Wu, Wan, Li, Tang, Zhang, and Zhuang]{yu2024anyedit}
Qifan Yu, Wei Chow, Zhongqi Yue, Kaihang Pan, Yang Wu, Xiaoyang Wan, Juncheng Li, Siliang Tang, Hanwang Zhang, and Yueting Zhuang.
\newblock Anyedit: Mastering unified high-quality image editing for any idea.
\newblock \emph{arXiv preprint arXiv:2411.15738}, 2024.

\bibitem[Zhang et~al.(2023)Zhang, Rao, and Agrawala]{zhang2023adding}
Lvmin Zhang, Anyi Rao, and Maneesh Agrawala.
\newblock Adding conditional control to text-to-image diffusion models.
\newblock In \emph{Proceedings of the IEEE/CVF international conference on computer vision}, pages 3836--3847, 2023.

\bibitem[Zhang et~al.(2024)Zhang, Song, Liu, Wang, Yu, Tang, Li, Tang, Hu, Pan, et~al.]{zhang2024ssr}
Yuxuan Zhang, Yiren Song, Jiaming Liu, Rui Wang, Jinpeng Yu, Hao Tang, Huaxia Li, Xu Tang, Yao Hu, Han Pan, et~al.
\newblock Ssr-encoder: Encoding selective subject representation for subject-driven generation.
\newblock In \emph{Proceedings of the IEEE/CVF Conference on Computer Vision and Pattern Recognition}, pages 8069--8078, 2024.

\end{thebibliography}
}
% \newpage
% \maketitlesupplementary
% \input{sec/appendix}

% todos
% 1. Related work单独加一段，与UniReal的对比
% 2. conclusion加一段limits discuss
% 3. fig2 标注：textual condition，visual condition，visual generation
% 4. 实验里，要加入CogvideoX在图像生成上和SD1.5，SDXL，FLUX的效果对比：理想结论是，CogvideoX在图像生成质量上明显不如专有的图像生成模型，但依然在各类图像生成任务上取得了更好或者comparable的结果，证明了视频基模预训练的有效性，验证了我们的motivation
\end{document}